# Utilizing AI and Machine Learning for Predictive Analysis of Post-Treatment Cancer Recurrence

[1]Muhammad Umer Qayyum, [2]Muhammad Fahad, [3]Nasrullah Abbasi.

[1,2,3]Washington University of Science and Technology, Virginia, USA

[1]qayyum.student@wust.edu, [2]fahad.student@wust.edu, [3]nabbasi.studnet@wust.edu,

## Abstract

In oncology, recurrence after treatment is one of the major challenges, related to patients' survival and quality of life. Conventionally, prediction of cancer relapse has always relied on clinical observation with statistical model support, which almost fails to explain the complex, multifactorial nature of tumor recurrence. This research explores how AI and ML models may increase the accuracy and reliability of recurrence prediction in cancer. Therefore, AI and ML create new opportunities not only for personalized medicine but also for proactive management of patients through analyzing large volumes of data on genetics, clinical manifestations, and treatment. The paper describes the various AI and ML techniques for pattern identification and outcome prediction in cancer patients using supervised and unsupervised learning. Clinical implications provide an opportunity to review how early interventions could happen and the design of treatment planning.

## Introduction

Cancer recurrence is a critical concern in oncology, with significant implications for patient prognosis and treatment strategies. Despite advancements in cancer treatment, recurrence remains a major challenge, often leading to reduced survival rates and diminished quality of life. Traditional methods of predicting cancer recurrence typically rely on clinical factors such as tumor stage, histological grade, and patient demographics. While these methods provide valuable insights, they often fall short in accurately predicting which patients are at the highest risk of relapse. The advent of AI and ML technologies presents a promising solution to this challenge. These technologies have the potential to analyze vast amounts of data, uncovering patterns and associations that may not be evident through traditional statistical methods. AI and ML models can integrate diverse data sources, including genomic data, imaging, and electronic health records, to develop personalized predictions for cancer recurrence. This research paper aims to explore the role of AI and ML in enhancing the predictive accuracy of post-treatment cancer recurrence, offering new opportunities for personalized patient care and early intervention.

### Purpose of the Study

- To explore the effectiveness of AI and ML models in predicting cancer recurrence after treatment.
- To identify key factors that contribute to cancer relapse using advanced data analytics.
- To evaluate the potential of AI and ML for personalized medicine, particularly in creating individualized risk assessments and treatment plans.
- To assess the integration of AI and ML tools into clinical practice for proactive patient management.

### Method of Study

This study employs a comprehensive literature review and analysis of existing AI and ML models used in predicting cancer recurrence. The research focuses on identifying and evaluating different AI and ML algorithms, such as supervised learning (e.g., decision trees, random forests, support vector machines) and unsupervised learning (e.g., clustering algorithms), in their application to oncology. Data sources include peer-reviewed journals, clinical trial reports, and case studies that demonstrate the application of AI and ML in predicting cancer relapse. The study also considers real-world examples of AI-driven predictive models currently being used or developed in clinical settings.

### Discussion and findings

*The Complexity of Predicting Cancer Recurrence*

Predicting cancer recurrence is a highly complex challenge in the field of oncology, influenced by a wide array of interrelated factors. The multifactorial nature of cancer recurrence means that it is not just a matter of whether a patient will experience a relapse, but how various biological, genetic, and environmental factors interact to contribute to this outcome.

*Tumor Biology*

One of the most critical factors in cancer recurrence is the biological behavior of the tumor itself. Tumors are highly heterogeneous, with variations in cell types, genetic mutations, and molecular pathways. These differences can significantly influence the tumor's aggressiveness, its response to treatment, and its likelihood of returning after initial therapy. For instance, certain genetic mutations may make a tumor more resistant to chemotherapy, leading to incomplete eradication and a higher chance of recurrence. Additionally, the tumor microenvironment, which includes surrounding blood vessels, immune cells, and other supportive tissue, can also play a crucial role in recurrence by providing a favorable environment for residual cancer cells to grow and spread (Rodríguez-Vicente et al., 2016).

i.   2. Patient Genetics

Patient genetics is another significant factor contributing to cancer recurrence. Genetic predispositions, such as BRCA1 and BRCA2 mutations in breast cancer, can significantly increase the risk of cancer returning. These genetic factors can influence how the patient's body metabolizes drugs, repairs DNA, and responds to treatment, all of which can affect the likelihood of recurrence. Moreover, genetic diversity among patients means that what works for one individual may not work for another, making it challenging to predict recurrence using a one-size-fits-all approach. (*BRCA Gene Mutations: Cancer Risk and Genetic Testing Fact Sheet*, 2024).

ii.   3. Treatment Modalities

The type and effectiveness of the treatment modalities used also play a crucial role in cancer recurrence. While treatments like surgery, radiation, and chemotherapy aim to eliminate cancer cells, they may not always be entirely successful. Residual microscopic disease, where small clusters of cancer cells remain undetected in the body, can lead to relapse. Moreover, cancer cells can develop resistance to treatment over time, particularly with chemotherapy and targeted therapies, which can contribute to recurrence. The timing, dosage, and combination of treatments can also affect the likelihood of recurrence, adding another layer of complexity to predictions. (*Radiation Therapy for Cancer*, 2019)

*Lifestyle Factors*

Lifestyle factors, such as diet, physical activity, smoking, and alcohol consumption, can influence cancer recurrence. (Molina-Montes et al., 2021). For example, obesity has been linked to an increased risk of recurrence in several cancers, including breast and colorectal cancer, possibly due to the inflammatory environment it creates. Additionally, stress, environmental exposures, and other lifestyle factors can affect the body's immune response, potentially allowing residual cancer cells to proliferate. These variables add another dimension to the prediction of cancer recurrence, making it necessary to consider them alongside biological and genetic factors

*Limitations of Traditional Statistical Models*

Traditional statistical models, such as Cox proportional hazards models, have been widely used in oncology to predict cancer recurrence based on clinical variables like tumor stage and patient age. However, these models often assume linear relationships between variables and may not capture the complex, non-linear interactions between multiple factors that influence recurrence. (Matsuo et al., 2019). Matsue et al. study showed the following result given in the Table.

**Table.1**

| Characteristic | Details |
|---|---|
| Total Number of Women Included | 768 |
| Median Age | 49 years |
| Ethnicity | Majority Hispanic (71.7%) |
| Tumor Type | Majority Squamous (75.3%) |
| Tumor Stage | Majority Stage I (48.7%) |
| Median Follow-up Time | 40.2 months |
| Number of Recurrence/Progression Events | 241 |
| Number of Deaths | 170 |
| Performance of Deep-Learning Model (PFS Prediction) | Mean Absolute Error: 29.3 |
| Performance of Cox Proportional Hazard Model (PFS) | Mean Absolute Error: 316.2 |
| Performance of Deep-Learning Model (Overall Survival) | Mean Absolute Error: 30.7 |
| Performance of Cox Proportional Hazard Model (OS) | Mean Absolute Error: 43.6 |
| Concordance Index for PFS with 20 Features | 0.695 |
| Concordance Index for PFS with 36 Features | 0.787 |
| Concordance Index for PFS with 40 Features | 0.795 |
| Significant Features (PFS) in Deep-Learning Model Only | 10 |
| Significant Features (OS) in Deep-Learning Model Only | 3 |
| Significant Features (PFS) in Cox Model Only | 0 |
| Significant Features (OS) in Cox Model Only | 3 |

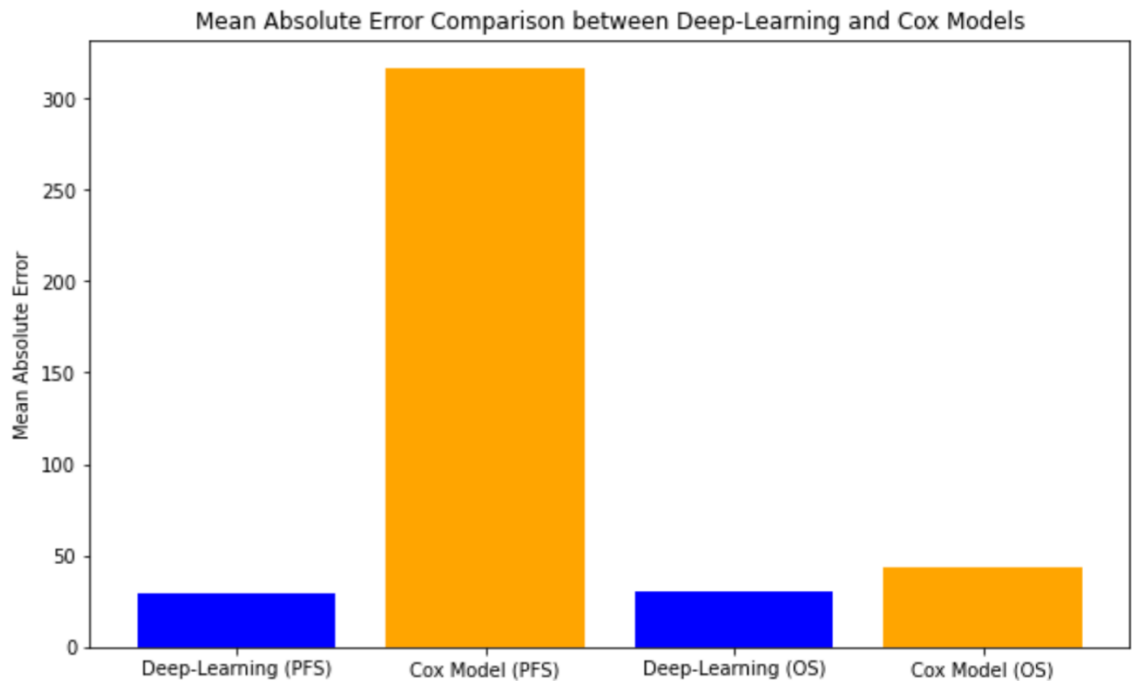

Data Source: https://www.ncbi.nlm.nih.gov/pmc/articles/PMC7526040/

*The Promise of AI and Machine Learning*

AI and machine learning (ML) offer a promising solution to these challenges by leveraging computational power to analyze large and complex datasets. Unlike traditional models, AI and ML algorithms can handle high-dimensional data and identify non-linear relationships between variables. This capability allows them to detect subtle patterns and interactions that might be missed by human analysts or simpler statistical models. For instance, AI algorithms can analyze genomic data to identify mutations associated with recurrence, or they can process imaging data to detect changes in tumor morphology that may signal a higher risk of relapse (Sebastian & Peter, 2022). A notable example of the application of ML in cancer recurrence prediction is the study by (Xu et al., 2020) which demonstrated the use of machine learning algorithms to analyze a combination of clinical data and gene expression profiles. The study found that ML models significantly outperformed traditional statistical methods in predicting patient outcomes, highlighting the potential of these technologies to provide more accurate and individualized predictions. (Vadapalli et al., 2022). Similarly, deep learning techniques were also utilized to analyze radiomic features extracted from imaging data. Radiomics involves the extraction of large amounts of quantitative features from medical images, which can then be used to create predictive models. The study achieved high accuracy in predicting recurrence in lung cancer patients, demonstrating the power of deep learning to process complex imaging data and improve predictive accuracy. These studies underscore the potential of AI and ML to revolutionize the field of oncology by providing more accurate and personalized predictions for cancer recurrence.

### *Integration into Clinical Practice*

The integration of AI and ML into clinical practice has the potential to transform how oncologists predict and manage cancer recurrence. By providing more accurate predictions, these technologies can enable earlier and more targeted interventions, improving patient outcomes. For example, patients identified as high-risk for recurrence could be monitored more closely, receive more aggressive treatment, or be enrolled in clinical trials for new therapies. Furthermore, AI-driven models can help oncologists personalize treatment plans based on an individual's unique risk profile, moving towards a more tailored approach to cancer care.

## AI and Machine Learning Techniques in Predictive Modeling

Artificial Intelligence (AI) and Machine Learning (ML) have revolutionized the field of predictive modeling, particularly in the context of cancer recurrence. These technologies encompass a broad spectrum of techniques, each with unique capabilities and applications in oncology. By leveraging these advanced models, researchers and clinicians can gain deeper insights into the factors driving cancer recurrence, enabling more accurate predictions and personalized treatment strategies.

### *Supervised Learning Models*

Supervised learning is one of the most commonly used AI techniques in cancer recurrence prediction. In supervised learning, models are trained on labeled datasets, where the outcome (such as recurrence or no recurrence) is already known. This allows the model to learn the associations between input variables (such as patient demographics, tumor characteristics, and treatment history) and the outcome, thereby enabling it to predict the likelihood of recurrence in new patients. Among the most widely used supervised learning models are decision trees, random forests, and support vector machines (SVMs).

**Decision Trees**: Decision trees operate by splitting the data into branches based on certain decision rules derived from the input features. Each branch represents a possible outcome, and by traversing the tree, the model can classify patients based on their recurrence risk. Decision trees are particularly valuable because they provide a clear and interpretable decision-making process, making them easier to understand and apply in clinical settings.

**Random Forests**: Random forests extend the decision tree concept by creating an ensemble of trees, each trained on a different subset of the data. The final prediction is made by aggregating the predictions from all the trees. This approach helps to reduce overfitting and improves the model's robustness and accuracy. In the context of cancer recurrence, random forests have been successfully applied to integrate diverse data types, such as clinical variables, genomic data, and imaging features, providing a comprehensive risk assessment.

**Support Vector Machines (SVMs)**: SVMs are powerful classifiers that work by finding the optimal hyperplane that separates the data into different classes. In cancer recurrence prediction, SVMs can handle high-dimensional data and are particularly effective in cases where the classes (recurrence and no recurrence) are not linearly separable. By applying kernel functions, SVMs can map the data into a higher-dimensional space, making it easier to find the separating hyperplane.

*Unsupervised Learning Techniques*

In contrast to supervised learning, unsupervised learning techniques do not rely on labeled data. Instead, these models aim to uncover hidden patterns or structures within the data by grouping patients based on similarities in their features. This approach is particularly valuable in cancer recurrence prediction, as it can identify previously unrecognized subgroups of patients with distinct risks, leading to more targeted treatment strategies.

**Clustering Algorithms**: Clustering algorithms, such as K-means, hierarchical clustering, and Gaussian mixture models, are commonly used in unsupervised learning to group patients into clusters based on their feature similarities. For example, K-means clustering partitions the data into a specified number of clusters, where each patient is assigned to the cluster with the nearest mean value of the features. In cancer recurrence, clustering can reveal subpopulations of patients who share similar genetic mutations, tumor characteristics, or treatment responses, enabling more personalized care.

**Dimensionality Reduction Techniques**: Techniques like Principal Component Analysis (PCA) and t-Distributed Stochastic Neighbor Embedding (t-SNE) are used to reduce the dimensionality of complex datasets while preserving their essential structure. These techniques can be particularly useful in visualizing high-dimensional data, such as genomic sequences or imaging features, and identifying patterns that may be associated with recurrence risk. By simplifying the data, these methods make it easier to apply other machine learning techniques and interpret the results.

*Deep Learning Models*

Deep learning, a subset of machine learning, has shown immense promise in predictive modeling for cancer recurrence, particularly in analyzing complex data types such as medical images, genomic data, and longitudinal patient records. Deep learning models, including Convolutional Neural Networks (CNNs) and Recurrent Neural Networks (RNNs), excel at learning hierarchical representations of data, which allows them to capture intricate patterns that may be predictive of recurrence.

**Convolutional Neural Networks (CNNs)**: CNNs are particularly effective in image analysis, making them a valuable tool for analyzing medical images, such as histopathological slides, radiographs, and MRI scans. CNNs consist of multiple layers that automatically learn to extract features from the images, such as edges, textures, and shapes, which are then used to make predictions. In cancer recurrence prediction, CNNs can analyze histopathological slides to detect subtle morphological changes in tumor cells that may indicate a higher risk of.

**Recurrent Neural Networks (RNNs)**: RNNs are designed to handle sequential data, making them well-suited for analyzing longitudinal patient data, such as changes in tumor markers or treatment responses over time. RNNs can capture temporal dependencies in the data, allowing them to make predictions based on the progression of the disease. In cancer recurrence prediction, RNNs can be used to monitor patient health over time and predict the likelihood of recurrence based on trends in clinical and laboratory data. In a notable study, (Zhu et al. 2022) applied deep learning to predict breast cancer recurrence by analyzing digitized histopathological slides. The deep learning model was able to identify subtle features associated with aggressive tumor behavior, providing valuable insights into recurrence risk. This study highlights the potential of deep learning to enhance the accuracy of cancer recurrence predictions by leveraging detailed image analysis, thereby offering a more nuanced understanding of the factors driving recurrence.

i. **Hybrid and Ensemble Models**

In addition to the individual models mentioned above, hybrid and ensemble approaches are increasingly being used in predictive modeling for cancer recurrence. These methods combine multiple models to leverage their respective strengths and improve overall predictive performance.

**Hybrid Models**: Hybrid models integrate different types of machine learning techniques, such as combining a CNN with an SVM, to create a more powerful predictive model. For instance, a hybrid model might use a CNN to extract features from medical images and then apply an SVM to classify patients based on those features. This approach can improve accuracy by combining the strengths of both models (Zhang & Wu, 2020).

**Ensemble Models**: Ensemble models, such as random forests and gradient boosting machines, combine the predictions of multiple models to achieve better performance. These models are particularly effective in reducing variance and improving robustness, as they aggregate the outputs of several models to make a final prediction. In cancer recurrence prediction, ensemble models have been shown to outperform individual models by providing more stable and accurate predictions (Dietterich, 2000).

*Interpretability and Explainability in AI Models*

One of the challenges in applying AI and ML techniques to cancer recurrence prediction is the interpretability of the models. While deep learning models, in particular, can achieve high predictive accuracy, they are often described as "black boxes" because their decision-making processes are not easily understood. This lack of transparency can be a barrier to clinical adoption, as clinicians may be hesitant to rely on models that they cannot fully interpret. To address this challenge, researchers are developing explainable AI (XAI) techniques that aim to make AI models more transparent and interpretable. For example, techniques such as SHAP (SHapley Additive exPlanations) and LIME (Local Interpretable Model-agnostic Explanations) are used to explain the contributions of individual features to the model's predictions. By providing clinicians with insights into how the model arrived at its prediction, XAI techniques can build trust in AI-driven decision-making and facilitate the integration of AI models into clinical practice (Ribeiro, Singh, & Guestrin, 2016).

2. **Data Integration and Personalized Medicine**

One of the key advantages of AI and ML in predicting cancer recurrence is the ability to integrate diverse data sources. Traditional models often rely on a limited number of clinical variables, but AI can incorporate data from genomics, proteomics, radiomics, and electronic health records. This integrative approach enables the development of more comprehensive models that consider the multifactorial nature of cancer (Gulshan et al., 2016). Genomic data can provide insights into the molecular mechanisms driving cancer recurrence, while imaging data can reveal changes in tumor morphology that may indicate a higher risk of relapse. By combining these data sources, AI and ML models can generate personalized risk assessments, guiding clinicians in tailoring treatment plans to the individual patient. This approach aligns with the principles of precision medicine, where treatment strategies are customized based on the patient's unique characteristics

## Clinical Implementation and Challenges

The integration of AI and machine learning (ML) into clinical practice for predicting cancer recurrence holds immense potential, yet it is fraught with significant challenges. These challenges must be carefully navigated to realize the full benefits of these advanced technologies in improving patient outcomes. The complexities of clinical implementation involve not only technical and logistical hurdles but also ethical, legal, and social considerations that must be addressed to ensure the safe and effective use of AI in healthcare.

### *Requirement for Large, High-Quality Datasets*

One of the most critical challenges in implementing AI and ML in cancer recurrence prediction is the need for large, high-quality datasets. Cancer is an inherently heterogeneous disease, with vast variations in tumor biology, patient genetics, treatment responses, and environmental influences. These variations make it challenging to develop AI models that can generalize well across diverse patient populations. The quality of the data used to train these models is paramount; it must be comprehensive, accurate, and representative of the broader patient population to avoid biases and ensure the model's applicability in real-world settings. However, acquiring such datasets is not without its difficulties. Many healthcare institutions operate in silos, with patient data scattered across different systems and formats. Data standardization is another critical issue, as inconsistencies in data entry, coding practices, and diagnostic criteria can lead to discrepancies that reduce the model's reliability. Additionally, the longitudinal nature of cancer recurrence prediction requires long-term follow-up data, which is often incomplete or unavailable due to patient attrition or changes in healthcare providers. Collaborative efforts to create large, standardized datasets, such as multi-institutional consortia and data-sharing initiatives, are essential to overcome these challenges and enhance the robustness of AI models (Gulshan et al., 2016).

### *Data Privacy and Security Concerns*

Data privacy and security are paramount when integrating AI with clinical data, especially when dealing with sensitive information such as genomic data, which can reveal deeply personal details about a patient's predisposition to certain diseases. The integration of such data into AI systems poses significant privacy risks, including potential breaches that could lead to unauthorized access, misuse of patient information, and loss of patient trust. Moreover, the legal landscape surrounding data privacy is complex and varies significantly across different regions. For instance, in the United States, the Health Insurance Portability and Accountability Act (HIPAA) sets stringent standards for the protection of patient information, while the European Union's General Data Protection Regulation (GDPR) imposes strict regulations on data processing and transfer. Navigating these legal frameworks is a complex task that requires careful consideration of compliance issues to ensure that AI systems are not only effective but also ethically and legally sound (He et al., 2019). To address these concerns, robust data encryption, anonymization, and secure data-sharing protocols must be implemented. Additionally, developing AI models that require minimal patient data, or that can function effectively with anonymized data, could further reduce privacy risks. Continuous monitoring and updating of data security measures are also necessary to keep pace with evolving cyber threats and ensure the ongoing protection of patient information.

*Model Interpretability and Transparency*

A significant challenge in the clinical implementation of AI and ML is the interpretability of the models. While AI models, particularly deep learning models, can achieve remarkable predictive accuracy, they are often criticized for being "black boxes." These models make decisions based on complex, often non-linear interactions between a multitude of variables, making it difficult for clinicians to understand the rationale behind a prediction. This lack of transparency can be a major barrier to clinical adoption, as healthcare providers may be reluctant to rely on predictions they cannot fully comprehend or explain to their patients (Topol, 2019). The development of explainable AI (XAI) models offers a promising solution to this challenge. XAI aims to create models that are not only accurate but also transparent and interpretable, allowing clinicians to understand how and why a particular decision was made. Techniques such as SHAP (SHapley Additive exPlanations) and LIME (Local Interpretable Model-agnostic Explanations) have been developed to provide insights into the contributions of individual features to the model's predictions, making the decision-making process more transparent (Ribeiro, Singh, & Guestrin, 2016). For instance, a study by Zhang and Wu (2020) demonstrated the use of XAI in predicting cancer recurrence. By providing visual explanations of the factors contributing to the prediction, the model helped clinicians better understand the underlying mechanisms driving the recurrence risk. This level of transparency is crucial for building trust in AI-driven decision-making and ensuring that these tools can be effectively integrated into clinical workflows.

*Integration into Clinical Workflows*

Another critical challenge is the seamless integration of AI models into existing clinical workflows. Healthcare providers are already burdened with a multitude of responsibilities, and the introduction of new technologies must be done in a way that complements rather than disrupts their practice. AI models must be user-friendly, providing actionable insights that clinicians can easily interpret and apply to patient care. The integration process also involves ensuring that AI systems are compatible with existing electronic health record (EHR) systems, which are often fragmented and vary significantly across institutions. Interoperability is a key concern, as AI models need to access and process data from different sources in real-time to provide accurate and timely predictions. Developing standardized interfaces and protocols for AI integration is essential to overcome these challenges and ensure that AI models can be effectively utilized in clinical settings (Esteva et al., 2017). Additionally, training and education are critical components of successful AI implementation. Clinicians need to be trained not only in the use of AI tools but also in understanding their limitations and potential biases. Continuous education programs and support from AI developers can help healthcare providers stay updated on the latest advancements and best practices in AI-driven healthcare.

*Future Directions and Research Opportunities*

The field of AI and ML in cancer recurrence prediction is rapidly evolving, with ongoing research aimed at addressing the challenges mentioned above. Future research should focus on developing explainable AI models that provide clinicians with insights into the factors driving the model's predictions. This transparency is crucial for building trust in AI-driven decision-making (Zhou & Pan, 2020). There is also a need for more research into the integration of AI models into clinical workflows. Developing user-friendly interfaces and decision support systems that seamlessly incorporate AI predictions into clinical practice will be essential for widespread adoption (Liu et al., 2019). Moreover, the potential of AI in drug development and treatment optimization presents an exciting avenue for future research. By predicting which patients are most likely to experience recurrence, AI can guide the development of targeted therapies that address the underlying causes of relapse. This approach could lead to the identification of new drug targets and the development of more effective treatment regimens (He et al., 2019).

**Conclusion**

The integration of Artificial Intelligence (AI) and Machine Learning (ML) in predicting cancer recurrence marks a transformative leap in the field of oncology. These technologies offer unprecedented potential to enhance patient outcomes by enabling early intervention and tailoring treatment strategies to the individual needs of each patient. Traditional methods of predicting cancer recurrence often rely on static models that may not fully capture the complex, multifactorial nature of cancer, leading to generalized treatment approaches. In contrast, AI and ML models can analyze vast and diverse datasets, including genomic data, imaging studies, and clinical records, to identify subtle patterns and interactions that are predictive of recurrence. Despite the immense promise, several challenges remain to be addressed for these technologies to be fully realized in clinical practice. High-quality, standardized data is essential for training robust models, yet obtaining such data can be difficult due to the heterogeneity of cancer and variations in patient records. Moreover, the "black box" nature of many AI models, particularly deep learning algorithms, poses issues of interpretability, making it challenging for clinicians to trust and adopt these tools without a clear understanding of their decision-making processes. Additionally, integrating AI into existing clinical workflows requires careful planning to ensure these tools complement rather than disrupt patient care. Nevertheless, the potential benefits are substantial. By harnessing AI and ML, oncologists can move beyond one-size-fits-all approaches, offering personalized treatment plans that not only improve survival rates but also enhance the quality of life for cancer patients. As research continues to evolve, overcoming these challenges will be crucial for unlocking the full potential of AI in oncology, ultimately leading to more precise, effective, and patient-centered cancer care.